\documentclass[unnumsec,webpdf,contemporary,large]{oup-authoring-template}

\usepackage{xcolor}
\usepackage{titlesec}
\usepackage{hyperref}
\hypersetup{
    colorlinks=true,
    allcolors=cyan
}
\usepackage[superscript]{cite}
\definecolor{noblue}{RGB}{0,82,147}

\titleformat{\section}
  {\normalfont\Large\bfseries\color{black}}{\thesection}{1em}{}
\titleformat{\subsection}
  {\normalfont\large\bfseries\color{noblue}}{\thesubsection}{1em}{}

\graphicspath{{Fig/}}

\begin{document}

\journaltitle{Neuro-Oncology}
\DOI{DOI HERE}
\copyrightyear{2025}
\pubyear{2025}
\access{Advance Access Publication Date: Day Month Year}
\appnotes{Original Article}

\firstpage{1}

\title[Improving Generalization of Deep Learning for Brain Metastases Segmentation Across Institutions]{Improving Generalization of Deep Learning for Brain Metastases Segmentation Across Institutions}

\author[1]{Yuchen Yang}
\author[1]{Shuangyang Zhong}
\author[1]{Haijun Yu}
\author[1]{Langcuomu Suo}
\author[1,2]{Hongbin Han}
\author[3]{Florian Putz}
\author[1,2$\ast$]{Yixing Huang}
\authormark{Yang et al.}

\address[1]{\orgdiv{Institute of Medical Technology}, \orgname{Peking University Health Science Center}, \orgaddress{Beijing, China}}
\address[2]{\orgdiv{Beijing Key Laboratory of Intelligent Neuromodulation and Brain Disorder Treatment},  \orgaddress{Beijing, China}}
\address[3]{\orgdiv{Department of Radiation Oncology}, \orgname{University Hospital Erlangen}, \orgaddress{Erlangen, Germany}}
\corresp[$\ast$]{Corresponding author: Yixing Huang, No. 38 Xueyuan Road, Haidian District, Beijing, China (\href{mailto:huangyx@pku.edu.cn}{huangyx@pku.edu.cn}).}


\abstract{\textbf{Background:} Deep learning has demonstrated significant potential for automated brain metastases (BM) segmentation; however, models trained at a singular institution often exhibit suboptimal performance at various sites due to disparities in scanner hardware, imaging protocols, and patient demographics. The goal of this work is to create a domain adaptation framework that will allow for BM segmentation to be used across multiple institutions.\\
\textbf{Methods:} We propose a VAE-MMD preprocessing pipeline that combines variational autoencoders (VAE) with maximum mean discrepancy (MMD) loss, incorporating skip connections and self-attention mechanisms alongside nnU-Net segmentation. The method was tested on 740 patients from four public databases: Stanford, UCSF, UCLM, and PKG, evaluated by domain classifier's accuracy, sensitivity, precision, F1/F2 scores, surface Dice (sDice), and 95th percentile Hausdorff distance (HD95) .\\
\textbf{Results:} VAE-MMD reduced domain classifier accuracy from 0.91 to 0.50, indicating successful feature alignment across institutions. Reconstructed volumes attained a PSNR greater than 36 dB, maintaining anatomical accuracy. The combined method raised the mean F1 by 11.1\% (0.700 to 0.778), the mean sDice by 7.93\% (0.7121 to 0.7686), and reduced the mean HD95 by 65.5\% (11.33 to 3.91 mm) across all four centers compared to the baseline nnU-Net.\\
\textbf{Conclusions:} VAE-MMD effectively diminishes cross-institutional data heterogeneity and enhances BM segmentation generalization—across volumetric, detection, and boundary-level metrics—without necessitating target-domain labels, thereby overcoming a significant obstacle to the clinical implementation of AI-assisted segmentation.}


\maketitle

\textbf{Importance of the Study}


Accurate segmentation of brain metastases (BM) is essential for stereotactic radiosurgery planning. Manual contouring, however,  takes a long time and can miss small, low-contrast lesions. Deep learning-based BM segmentation can make things faster and more sensitive, but models trained at just one institution often don't work well at other institutions because of scanner characteristics, imaging protocols, and patient populations. To tackle this issue, we present an innovative unsupervised domain adaptation framework that combines variational autoencoders with maximum mean discrepancy loss, augmented by skip connections and self-attention to maintain anatomical detail while acquiring domain-invariant features. We tested our method on 740 patients from four institutions (Stanford, UCSF, UCLM, and PKG). It lowered domain classifier accuracy from 91\% to 50\%, raised mean F1 by 11.1\%, improved mean sDice by 7.93\%, and reduced mean HD95 by 65.5\% (11.33$\to$3.91 mm) without requiring target-domain labels.

\section{Introduction}

Brain metastases (BM) are the most prevalent intracranial malignancy in adults, with an estimated incidence surpassing 100,000 new cases annually in the United States alone \cite{nayak2012epidemiology}. Up to 30\% to 40\% of patients with metastatic cancer develop brain metastases during their disease course, with lung cancer, melanoma, and breast cancer being the most frequent primary tumor sources \cite{barnholtz2004incidence,ostrom2018brain}. The incidence of BM has been steadily increasing, driven by improved diagnostic imaging sensitivity and by the development of systemic therapies that extend overall survival while the blood-brain barrier limits central nervous system drug penetration.\cite{arvanitis2020blood}

Stereotactic radiosurgery (SRS) is now the standard treatment for patients with limited BM. It has similar survival rates to whole-brain radiation therapy but with much less neurocognitive toxicity.\cite{brown2016effect,chang2009neurocognition,yamamoto2014srs} To deliver SRS effectively, radiation oncologists and neuroradiologists must manually contour the tumor's edges, correctly count the lesions, and measure the volume in detail.\cite{luo2024automated} However, manual segmentation is time-consuming and labor-intensive, and suffers from substantial inter-observer variation, especially for small lesions that may only be visible on a few image slices.\cite{lu2021randomized,ozkara2023deep}

Deep learning has shown great promise for automatically separating brain tumors, with architectures like U-Net \cite{ronneberger2015unet}, nnU-Net \cite{isensee2021nnunet}, and DeepMedic \cite{kamnitsas2017efficient} reaching expert-level performance on benchmark datasets. The nnU-Net framework \cite{isensee2021nnunet} has become the gold standard for medical image segmentation because it can automatically change preprocessing, architecture, and postprocessing to fit each dataset.\cite{isensee2021nnunet} Beyond single-timepoint segmentation, recent work has extended deep learning approaches to longitudinal tracking of BM, enabling automated detection and analysis of disease progression across serial MRI studies \cite{huang2022deep,machura2024deep}.  Clinical validation studies have shown that AI-assisted segmentation can improve both accuracy and efficiency. For instance, randomized multi-reader evaluations have demonstrated that automated segmentation tools can significantly reduce the clinician time (by up to 42\%) and facilitate consensus among different readers regarding their observations.\cite{luo2024automated,lu2021randomized}

Despite these advances, a major obstacle remains before clinical adoption: deep learning models trained at one institution often fail to generalize to data from other centers \cite{huang2024multicenter}. This phenomenon, referred to as domain shift or domain gap \cite{guan2022domain}, results from variability in imaging equipment, acquisition protocols, patient demographics, and annotation methodologies among institutions. Our prior research \cite{huang2024multicenter} thoroughly delineated domain gaps within six brain metastasis datasets, uncovering significant performance variability among centers. For example, the F1 scores ranged from 0.625 on the New York University (NYU) dataset to 0.876 on the University Hospital Erlangen (UKER) dataset. More importantly, our analysis showed that data diversity is a greater problem than data scarcity. Mixed multicenter training enhanced performance at specific institutions, notably Stanford University and New York University (NYU), while yielding minimal advantages at others, including University Hospital Zurich (USZ), the University of California, San Francisco (UCSF), and the Brain Tumor Segmentation Challenge (BraTS) datasets. These results show that simply combining data from different centers is insufficient to overcome domain shift.

\begin{figure*}[!t]
\centering
\begin{tabular}{cccc}
\includegraphics[width=0.24\textwidth]{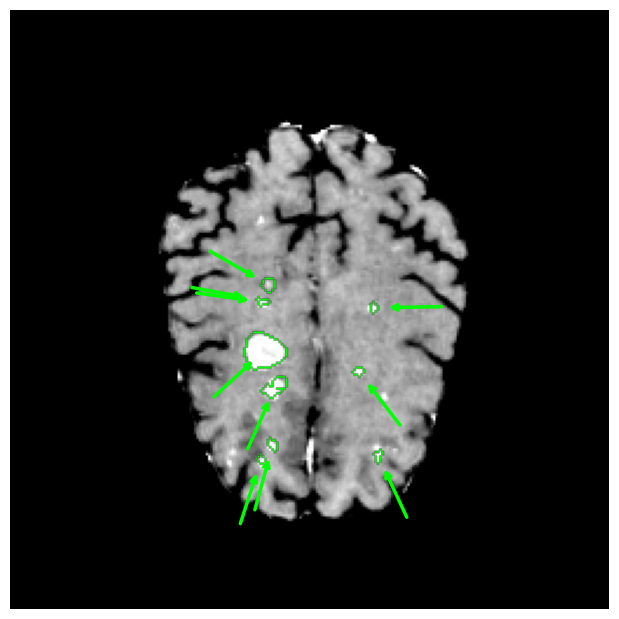} &
\includegraphics[width=0.24\textwidth]{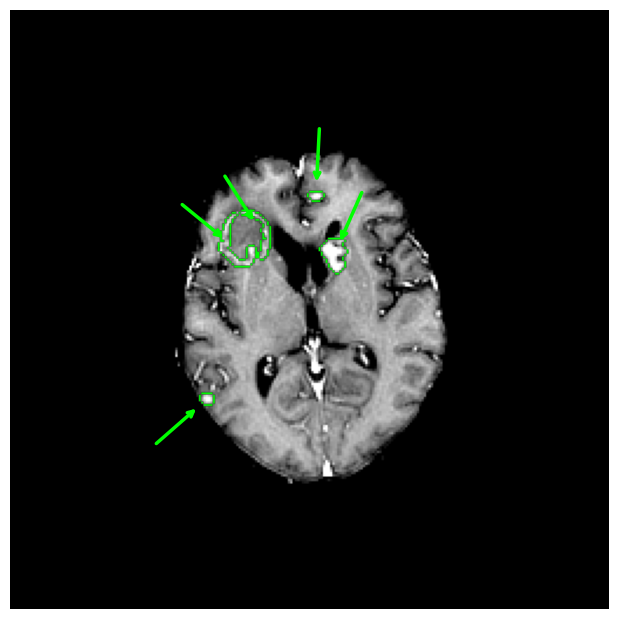} &
\includegraphics[width=0.24\textwidth]{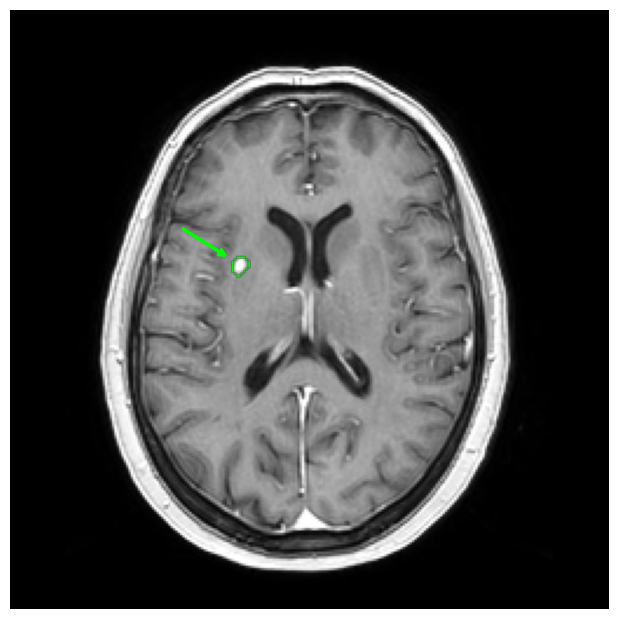} &
\includegraphics[width=0.24\textwidth]{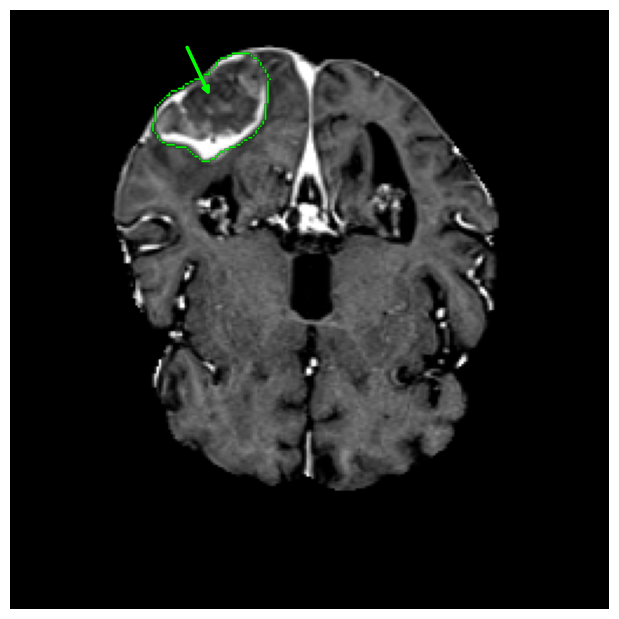} \\
(a) Stanford & (b) UCSF & (c) UCLM & (d) PKG
\end{tabular}
\caption{Representative axial T1-weighted post-contrast MRI slices from four institutions demonstrating heterogeneity in lesion characteristics. Green contours indicate expert-delineated metastases. (a) Stanford: multiple small scattered metastases characteristic of miliary disease pattern. (b) UCSF: mixed lesion sizes with prominent vascular enhancement. (c) UCLM: solitary well-circumscribed metastasis. (d) PKG: large solitary tumor from a patient with primary lung cancer. These differences in lesion burden and tumor morphology contribute to the domain gap that challenges cross-institutional model generalization.}
\label{fig:domain_gap}
\end{figure*}

The domain gap is evident in various dimensions, as demonstrated by representative cases from four institutions in Figure~\ref{fig:domain_gap}. The Stanford dataset (Fig. ~\ref{fig:domain_gap}a) shows a typical miliary pattern with many small enhancing lesions spread throughout the brain parenchyma. These lesions are hard to detect because of their small size and subtle appearance \cite{grovik2019deep}. The UCSF dataset (Fig. ~\ref{fig:domain_gap}b) shows a mixed picture with both bigger enhancing masses and smaller peripheral metastases. There is also a lot of vascular enhancement, which can make it hard for automated segmentation to work properly \cite{rudie2024ucsf}. The University of Castilla-La Mancha (UCLM) dataset (Fig. ~\ref{fig:domain_gap}c) usually shows solitary or oligometastatic disease with clear lesion edges \cite{ocana2023uclm}. The PKG dataset (Fig. ~\ref{fig:domain_gap}d) comprises patients with BM from primary lung cancer. Derived from surgical specimens with matched histopathology, the dataset often contains large resectable tumors. The domain shift in feature space can be clearly observed. Figure~\ref{fig:tsne_combined}(a) shows that the t-SNE visualization of nnU-Net features exhibits clear clustering by institution, with samples distinctly separated across centers. Most concerning, naive transfer learning methods result in catastrophic forgetting, with precision dropping from 0.900 to 0.418 when models were moved between centers. This indicates that models learn center-specific biases instead of generalizable anatomical knowledge.\cite{huang2024multicenter}

Domain adaptation techniques have surfaced as viable solutions to tackle data heterogeneity in medical imaging \cite{guan2022domain,kumari2024deep}. These methods aim to learn representations that are invariant to domain-specific features while preserving task-relevant information. Maximum mean discrepancy (MMD) \cite{gretton2012kernel}, first introduced as a kernel-based statistical test to compare distributions, is now a key tool for domain adaptation. MMD measures the distance between two probability distributions by comparing their embeddings in a reproducing kernel Hilbert space (RKHS), providing a principled way to make the source and target domains more alike. Variational autoencoders (VAEs) offer enhanced functionalities by obtaining compact latent representations that can be constrained to conform to specific distributions.\cite{kingma2014auto} VAE-based representation learning combined with MMD-based domain alignment enables the acquisition of features that are not confined to a specific domain, while preserving critical anatomical information for segmentation tasks.\cite{long2015learning}

In this research, we introduce an innovative domain adaptation framework for BM segmentation that amalgamates VAE-based feature learning with MMD-based domain alignment, integrated with nnU-Net segmentation. Our approach tackles three main problems: (1) learning representations that work across domains and narrow the measurable domain gap between institutions; (2) employing a composite reconstruction loss to keep anatomical information that is important for accurate segmentation; and (3) improving cross-institutional generalization without needing labeled data from target domains. We assess our methodology using four multi-institutional datasets, which include more than 700 patients from Stanford, UCSF, UCLM, and the PKG Brain-Mets-Lung-MRI collection, showcasing substantial enhancements in segmentation performance across all centers.

\section{Materials and Methods}

\subsection{Datasets}
This research employed four multi-institutional brain metastasis datasets from Stanford, UCSF, UCLM, and PKG, encompassing a total of 740 patients. These four datasets are available to the public and were released with the necessary Institutional Review Board (IRB) approvals and Data Transfer Agreements (DTAs). No internal or proprietary datasets were utilized, guaranteeing complete reproducibility of the study by other researchers.

The Stanford dataset, sourced from the Stanford BrainMetShare collection, comprised 105 patients with brain metastases originating from diverse primary malignancies.\cite{grovik2019deep} Each patient underwent co-registered T1-weighted post-gadolinium (T1CE) and FLAIR sequences, accompanied by expert-defined segmentations. This dataset contained numerous small lesions, with an average of 12.2 metastases per volume and 67\% of lesions measuring 0.1 cm$^3$ or less.

The UCSF dataset included 323 patients from the University of California San Francisco Bone Marrow Registry.\cite{rudie2024ucsf} The dataset comprised T1CE and FLAIR sequences along with their respective manual segmentations. UCSF cases typically had fewer but larger lesions and showed clear patterns of vascular enhancement, which was different from Stanford cases.

The UCLM dataset comprised 209 patients from the University Hospital of Ciudad Real, Spain.\cite{ocana2023uclm} This group included T1CE sequences only. Experienced neuroradiologists performed segmentation using semi-automated methods with manual refinement.

The PKG Brain-Mets-Lung-MRI dataset, which can be found on The Cancer Imaging Archive (TCIA), had 103 patients with brain metastases from primary lung cancer.\cite{pkg2023tcia} This dataset offered T1CE sequences along with their corresponding segmentations, representing a unique patient cohort characterized by homogeneous primary tumor histology.

We only used T1-weighted contrast-enhanced (T1CE) sequences for all datasets as they are the primary imaging modality for radiotherapy treatment planning and ensure consistency across centers. Each dataset was partitioned into a training set (80\%) and a test set (20\%). No patient overlap occurred between the two sets to ensure data independence. Five-fold cross-validation was applied to break the training set into smaller parts.


\subsection{Preprocessing}

The nnU-Net framework standardized the preprocessing of all images.\cite{isensee2021nnunet} First, images were resampled to 1 mm isotropic voxel spacing and cropped to a patch size of 64 x 192 x 160 voxels. To make sure that the spatial consistency was the same across datasets with different native resolutions, trilinear interpolation was used for images and nearest-neighbor interpolation was used for segmentation masks. We used z-score normalization based on foreground voxel statistics to normalize the intensity. Subsequently, the images were cut down to the non-zero area to reduce computational cost while keeping all the important anatomy.

For VAE training, preprocessed volumes were resized to a fixed size of 128 × 128 × 128 voxels using trilinear interpolation with a 1 mm isotropic voxel spacing. To match the tanh output activation of the VAE decoder, the intensity values were normalized to the range $[-1, 1]$ using min-max normalization. Data augmentation included random 3D flips along all three axes (with a 50\% probability for each) and random 90-degree rotations (with a 30\% probability).

\subsection{VAE-MMD Architecture}

\begin{figure*}[!t]
\centering
\begin{tabular}{cc}
\textbf{(a) VAE-MMD Architecture} & \textbf{(b) VAE-MMD Preprocessing for nnU-Net} \\[4pt]
\includegraphics[width=0.55\textwidth]{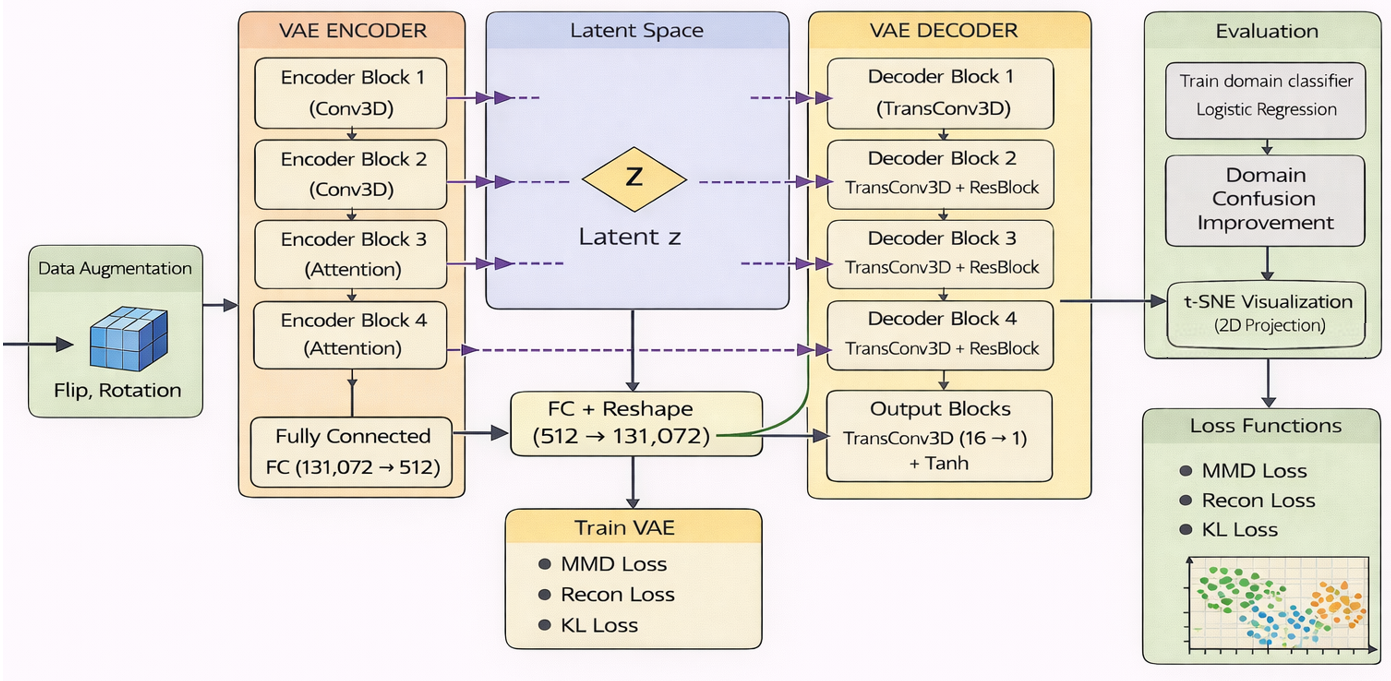} &
\includegraphics[width=0.40\textwidth]{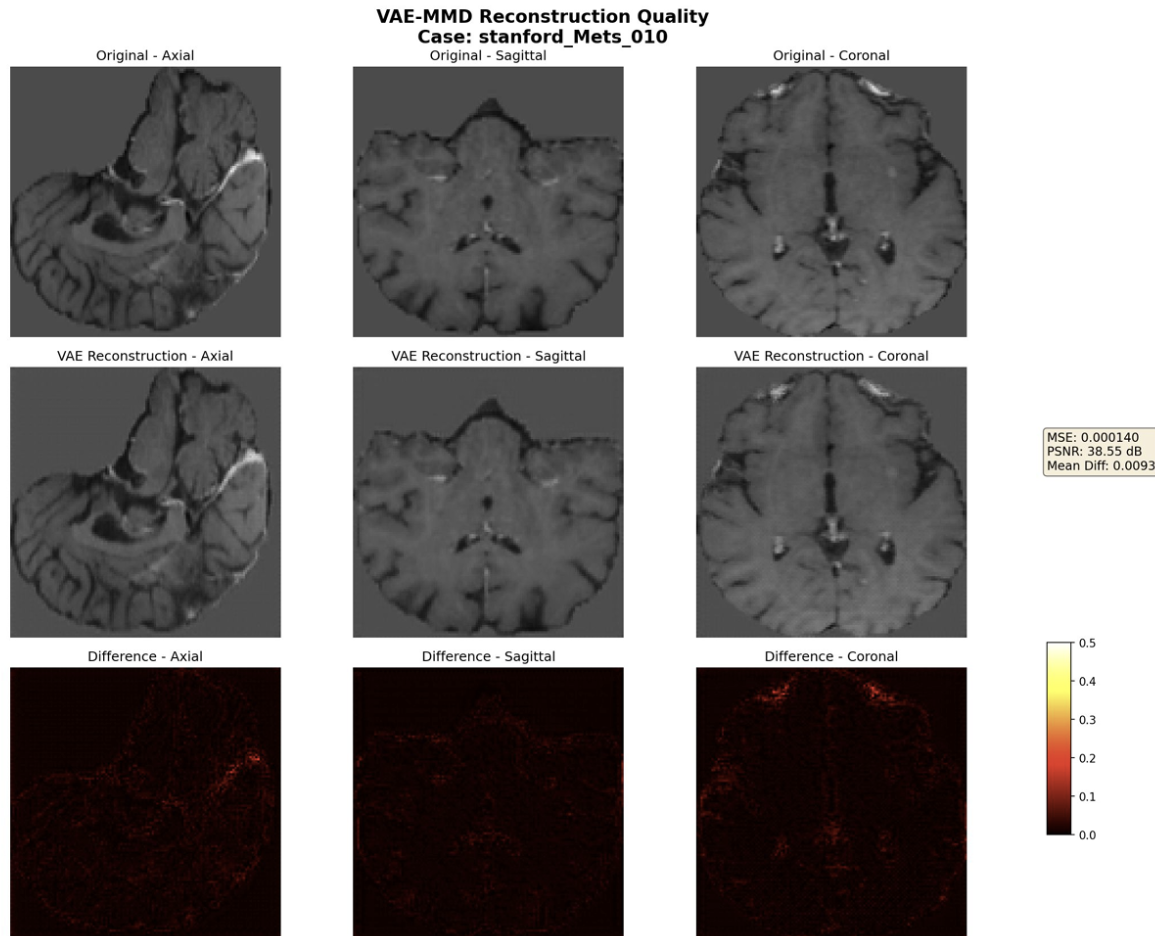} \\
\end{tabular}
\caption{VAE-MMD architecture and preprocessing pipeline. (a) The encoder progressively downsamples input volumes through four convolutional blocks (1$\rightarrow$32$\rightarrow$64$\rightarrow$128$\rightarrow$256 channels) with residual connections and self-attention modules. The latent space (512 dimensions) is regularized by KL divergence and aligned across domains via MMD loss. The decoder reconstructs images using transposed convolutions with skip connections from corresponding encoder blocks. (b) VAE-MMD reconstruction for a Stanford case (stanford\_Mets\_010): original images (top row) are passed through the VAE encoder-decoder to produce reconstructed volumes (middle row) in which institution-specific intensity distributions and scanner-related stylistic variations are normalized while lesion contrast and anatomical structures are preserved (PSNR = 38.55 dB, MSE = 0.000140). The near-black difference maps (bottom row, scale 0--0.5) confirm minimal information loss. These domain-harmonized reconstructions serve directly as nnU-Net inputs, reducing cross-institutional domain shift without requiring any target-domain labels.}
\label{fig:architecture}
\end{figure*}

We developed an enhanced VAE architecture that includes skip connections and self-attention mechanisms for learning features that work across different domains (Fig. \ref{fig:architecture}). The architecture was specifically designed to meet two conflicting goals: (1) learning domain-invariant representations through latent space alignment, and (2) keeping important anatomical details that are needed for accurate lesion reconstruction.

\subsubsection{Encoder}

The encoder consisted of four convolutional blocks that progressively decreased the input volume from $128^3$ to $8^3$, simultaneously expanding the channel dimension from 1 to 256 (1→32→64→128→256). This gradual channel expansion compensated for the loss of spatial information during downsampling by enabling each stage to represent more information.\cite{ronneberger2015unet}

A 3D convolution with a kernel size of $4 \times 4 \times 4$ and a stride of 2 made up each encoder block. This was followed by batch normalization, ReLU activation, and a residual block. The residual blocks, featuring two $3 \times 3 \times 3$ convolutions along with skip connections, addressed the issue of vanishing gradients in deep networks. This design allowed the model to effectively learn identity mappings when no further transformations were necessary.Dropout (rate=0.1) applied within residual blocks served as a regularization method to prevent overfitting on the limited training data.

Using query-key-value projections with a channel reduction factor of 8, encoder blocks 3 and 4 incorporated 3D self-attention modules. Self-attention was essential for capturing long-range spatial dependencies that convolutional layers alone struggle to model effectively, such as the connections between bilateral lesions or between a lesion and far-off anatomical landmarks.\cite{vaswani2017attention} We only incorporated attention modules in the deeper layers, where the feature maps were reduced in size to $16^3$ and $8^3$, as applying attention to high-resolution feature maps would be prohibitively costly.

\subsubsection{Latent Space}

The bottleneck layer flattened the $256 \times 8 \times 8 \times 8$ feature map to 131,072 dimensions. These were then projected through two fully connected layers to make the mean ($\mu$) and log-variance ($\log \sigma^2$) vectors, which each had 512 dimensions. This 512-dimensional latent space was approximately 4,000 times smaller than the original $128^3$ input. Consequently, the model needed to acquire compact, semantically rich representations rather than merely memorizing pixel-level details.

We employed the reparameterization trick to obtain latent samples: $z = \mu + \sigma \odot \epsilon$, where $\epsilon \sim \mathcal{N}(0, I)$.\cite{kingma2014auto} This formulation enabled backpropagation through the stochastic sampling operation, which made end-to-end training possible. The MMD loss function operated within this concealed space, aligning the distributions of samples from various institutions, thereby enabling their application across any domain.

\subsubsection{Decoder with Skip Connections}

The decoder's architecture was symmetric to that of the encoder, but employing transposed convolutions to upsample. It was very important that the skip connections from each encoder block were combined with the outputs from the corresponding decoder block. These skip connections had two main uses: (1) they provided high-frequency spatial details that would have been lost through the information bottleneck, enabling accurate reconstruction of fine structures such as small metastases; and (2) they created shorter gradient paths during backpropagation facilitating training of the deep network.\cite{ronneberger2015unet}

The decoder would have to use only the 512-dimensional latent vector to recreate all the spatial details without skip connections. This approach performed poorly in our initial experiments as it failed to keep the lesion boundaries. By putting together encoder features and decoder features at each resolution level, the network was able to combine domain‑invariant semantic information from the latent space with domain-specific spatial details from the encoder that were important for anatomy.

The last layer used a $1 \times 1 \times 1$ convolution to produce a single-channel output, which was then activated by tanh to constrain the output values to the range $[-1, 1]$, matching the input normalization range. The use of self-attention modules, skip connections, and composite reconstruction loss enabled high‑fidelity image reconstruction. All of the reconstructed volumes achieved a peak signal-to-noise ratio (PSNR) of more than 36 dB, indicating that our architecture preserved fine anatomical details while successfully adapting to the domain.

\subsection{Loss Functions}

The total VAE-MMD loss had four parts, each designed to achieve a specific objective in producing high-quality domain-invariant representations:
\begin{equation}
\mathcal{L}_{\text{total}} = \mathcal{L}_{\text{recon}} + \lambda_{\text{KL}} \mathcal{L}_{\text{KL}} + \lambda_{\text{MMD}} \mathcal{L}_{\text{MMD}} + \lambda_{\text{adv}} \mathcal{L}_{\text{adv}},
\end{equation}
where $\mathcal{L}_{\text{recon}}$, $\mathcal{L}_{\text{KL}}$, $\mathcal{L}_{\text{MMD}}$, and $\mathcal{L}_{\text{adv}}$ represent the reconstruction loss, Kullback-Leibler (KL) divergence loss, MMD loss, and adversarial loss, respectively. The relative contributions of the loss terms are governed by the weighting coefficients $\lambda_{\text{KL}}$, $\lambda_{\text{MMD}}$, and $\lambda_{\text{adv}}$.

\subsubsection{Reconstruction Loss}

The reconstruction loss ensured that the VAE latent space retained sufficient anatomical information for precise image reconstruction. This prevented the encoder from converging to trivial solutions, in which MMD could be minimized without preserving essential features. We combined three terms that work well together, as is common in neural network-based image reconstruction \cite{zhao2017loss}:
\begin{equation}
\mathcal{L}_{\text{recon}} = \lambda_{\text{L2}} \cdot \text{L2}(x, \hat{x}) + \lambda_{\text{L1}} \cdot \text{L1}(x, \hat{x}) + \lambda_{\text{SSIM}} \cdot (1 - \text{SSIM}(x, \hat{x})).
\end{equation}
The L2 term squares the penalties for large intensity deviations, effectively ensuring overall voxel‑wise accuracy and heavily penalizing outlier errors. The L1 term helps keep sharp edges and mitigates the over-smoothing that often happens with L2-only optimization, which is important for keeping the edges of lesions.\cite{zhao2017loss} The structural similarity index (SSIM) term preserved perceptual quality by comparing local patterns of luminance, contrast, and structure, ensuring that reconstructed images maintained anatomically meaningful textures rather than just matching pixel intensities.\cite{wang2004image}

We set $\lambda_{\text{L2}} = 300$, $\lambda_{\text{L1}} = 150$, and $\lambda_{\text{SSIM}} = 50$ to give more weight to accuracy of intensity while keeping structural fidelity. The higher L2 weight reflects its importance in medical imaging, where absolute intensity values are important for diagnosis. This combination of losses has been shown to outperform single loss functions in MRI reconstruction tasks.\cite{zhao2017loss}

\subsubsection{KL Divergence}
The KL divergence regularized the latent distribution toward a standard Gaussian:
\begin{equation}
\mathcal{L}_{\text{KL}} = -\frac{1}{2} \sum_{j=1}^{512} \left(1 + \log \sigma_j^2 - \mu_j^2 - \sigma_j^2\right),
\end{equation}
where $\mu_j$ and $\sigma_j^2$ stand for the mean and variance of the $j$-th dimension of the encoded latent distribution.
This regularization served two main purposes: (1) preventing the latent space from fragmenting into separate clusters that would hinder align domains, and (2) ensuring a smooth and continuous latent manifold, thereby enabling meaningful interpolation between samples.\cite{kingma2014auto} We set $\lambda_{\text{KL}} = 0.1$ to impose a moderate regularization on the latent space without introducing excessive tightness. This low weight followed the $\beta$-VAE framework, which suggests that lower KL weighting keeps the quality of the reconstruction while keeping the structure of the latent space.\cite{higgins2017beta}

\subsubsection{MMD Loss}

The maximum mean discrepancy loss was the core component for domain adaptation. It reduced the distance between institutions in the latent space:\cite{gretton2012kernel}
\begin{multline}
\mathcal{L}_{\text{MMD}} = \frac{1}{|\sigma|} \sum_{\sigma \in \{0.5, 1.0, 2.0\}} \Big[ \mathbb{E}[k_\sigma(z^s, z^{s'})] \\
+ \mathbb{E}[k_\sigma(z^t, z^{t'})] - 2\mathbb{E}[k_\sigma(z^s, z^t)] \Big],
\end{multline}
where $k_\sigma(z, z') = \exp(-\|z - z'\|^2 / 2\sigma^2)$ is the kernel for the radial basis function. We used multi-scale kernels ($\sigma \in \{0.5, 1.0, 2.0\}$) to capture domain differences at varying levels of detail. Smaller bandwidths captured fine‑grained distributional differences, while larger bandwidths captured global distributional shifts.\cite{long2015learning} The weight $\lambda_{\text{MMD}} = 10$ was set relatively high compared to KL divergence to make domain alignment the main goal, in addition to reconstruction quality.

\subsubsection{Adversarial Loss}

A 3D discriminator network provided adversarial training to improve reconstruction realism:
\begin{equation}
\mathcal{L}_{\text{adv}} = \mathbb{E}[(D(\hat{x}) - 1)^2]
\end{equation}
The adversarial loss enabled the VAE to produce reconstructions that closely resembled authentic images, effectively eliminating artifacts and enhancing perceptual quality beyond what pixel-wise losses could achieve independently. We established $\lambda_{\text{adv}} = 5$ to provide a beneficial gradient signal while preventing training from becoming unstable due to mode collapse.

\subsection{Training Protocol}

We used the Adam optimizer with a learning rate of $1 \times 10^{-4}$, $\beta_1 = 0.9$, $\beta_2 = 0.999$, and a batch size of 4 to train the VAE-MMD model. Training lasted for 100 epochs, and to keep it stable, the gradient was clipped (max norm = 1.0). The identical optimizer settings were employed to train the discriminator at various intervals.

Each training iteration incorporated samples from all four domains within the batches. We computed the MMD loss for every pair of domains within the batch and subsequently calculated their average. The optimal model was selected by identifying the one with the lowest total VAE loss on a validation set that had not been utilized during the training process.

\subsection{Segmentation with nnU-Net}

We employed the nnU-Net framework in its 3D full-resolution configuration for the subsequent segmentation task.\cite{isensee2021nnunet} The framework employed dataset fingerprint analysis to autonomously identify the optimal preprocessing parameters, network architecture, and training hyperparameters. We employed 5-fold cross-validation utilizing the integrated multi-institutional training dataset to develop the model. To evaluate the impact of domain adaptation on nnU-Net, it was trained using both the original images (baseline) and those reconstructed through a VAE (VAE-MMD preprocessing).

\subsection{Evaluation Metrics}

We evaluated the effectiveness of domain adaptation by training a logistic regression classifier to utilize VAE latent features for predicting the institution label. A decrease in the accuracy of the lower domain classifier indicated a stronger alignment among the domains.

We employed the Dice similarity coefficient (DSC) to assess voxel-wise overlap, while the lesion-wise F1 score was utilized to evaluate detection accuracy, thereby gauging the effectiveness of the segmentation process. A lesion was considered detected if any predicted voxel overlapped with the ground truth. At the lesion level, we determined the precision and sensitivity. We additionally employed surface Dice (sDice) with a tolerance of 1 mm, along with the 95th percentile Hausdorff distance (HD95), to evaluate the quality of surface-level segmentation. These metrics were derived from binary surface maps generated through morphological erosion. To reduce the sensitivity of HD95 to outliers, the median was employed to summarize it across the various test cases. To evaluate the model performance across different institutions, we computed all metrics individually for the test set of each institution.

\section{Results}

\subsection{Domain Adaptation Effectiveness}

Initially, we examined the latent space representations alongside the efficacy of the domain classifier to determine whether VAE-MMD effectively bridged the divide between institutions.

\subsubsection{Two-Center Validation (Stanford and UCSF)}

\begin{figure*}[!t]
\centering
\begin{tabular}{cc}
\multicolumn{2}{c}{\textbf{(a) t-SNE Visualization}} \\[2pt]
\includegraphics[width=0.42\textwidth]{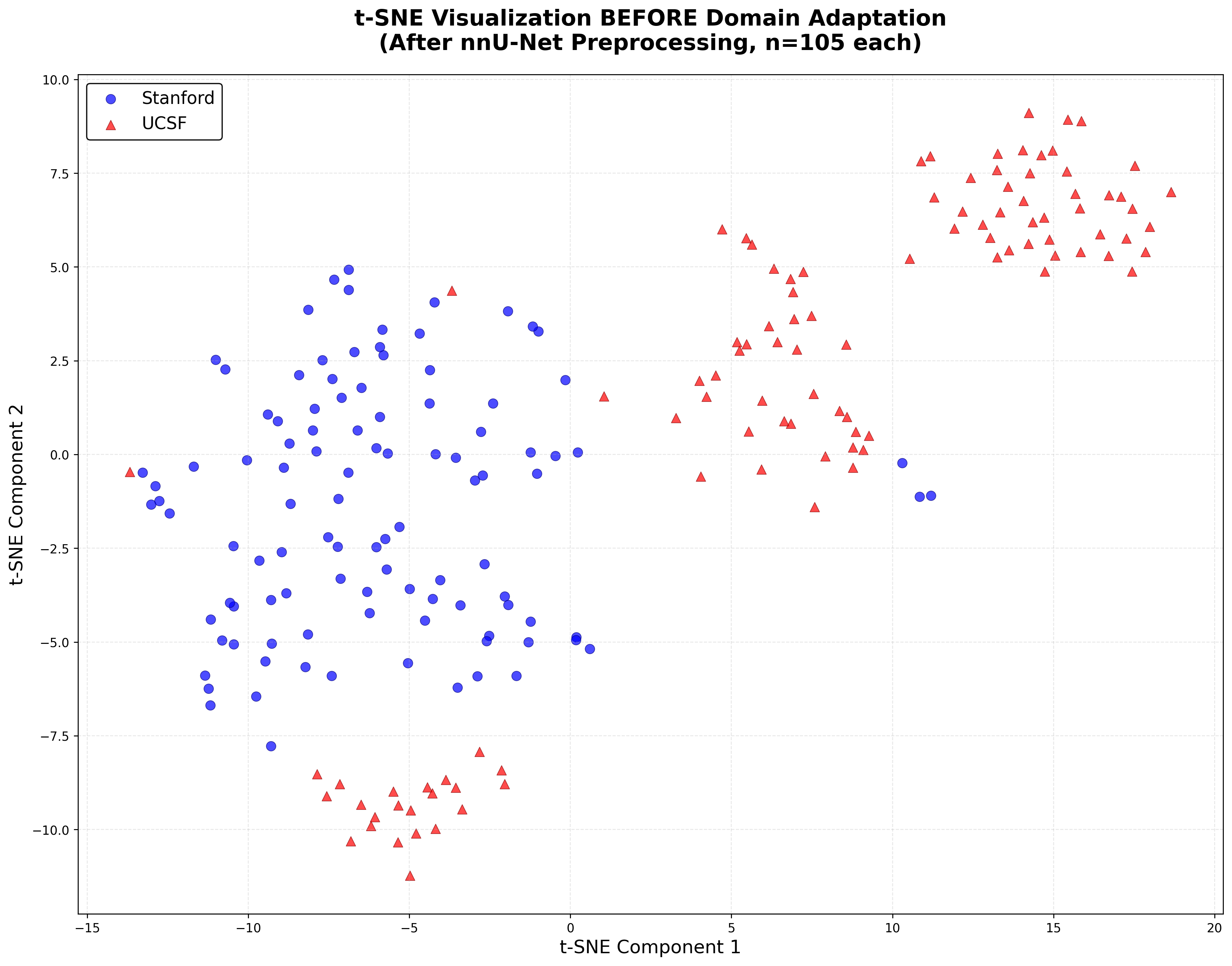} &
\includegraphics[width=0.42\textwidth]{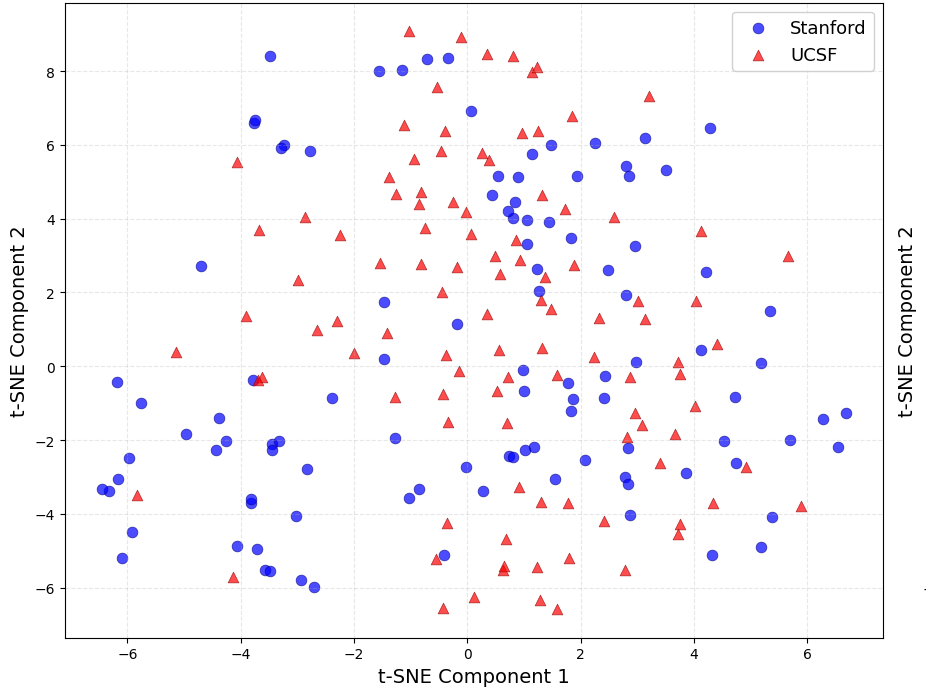} \\
{\small Before VAE-MMD (2-center)} & {\small After VAE-MMD (2-center)} \\[6pt]
\multicolumn{2}{c}{\includegraphics[width=0.85\textwidth]{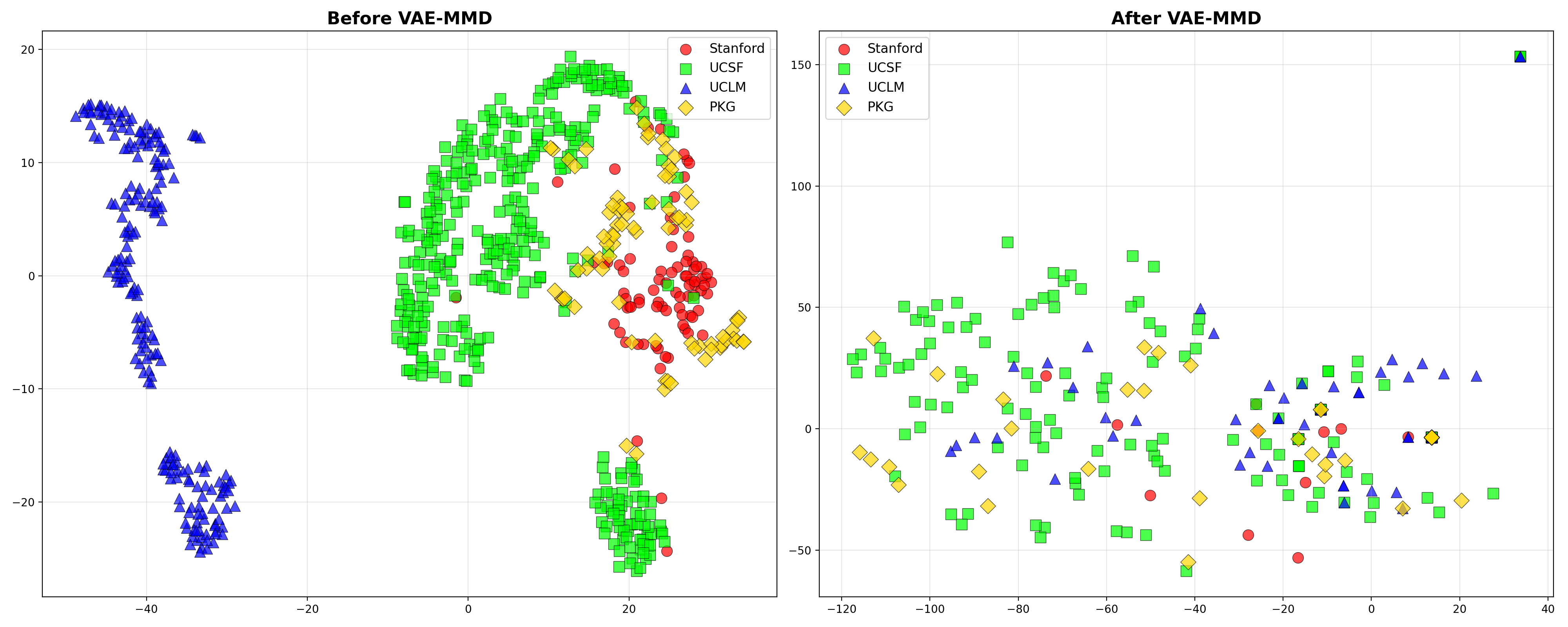}} \\
\multicolumn{2}{c}{\small Before VAE-MMD (4-center) \hspace{2cm} After VAE-MMD (4-center)} \\[8pt]
\multicolumn{2}{c}{\textbf{(b) Domain Classifier Confusion Matrices}} \\[2pt]
\multicolumn{2}{c}{\includegraphics[width=0.85\textwidth]{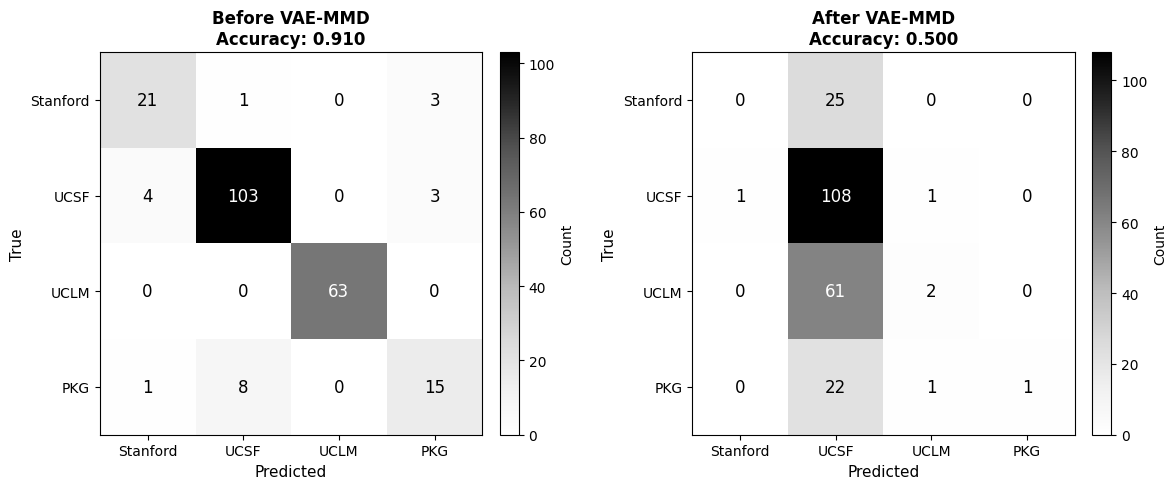}} \\
\end{tabular}
\caption{Domain adaptation evaluation. (a) t-SNE visualization: two-center (top) and four-center (bottom) configurations before and after VAE-MMD, showing progressive inter-institutional mixing. (b) Confusion matrices before and after VAE-MMD: accuracy declines from 91.0\% to 50.0\%, confirming elimination of institution-specific feature signatures.}
\label{fig:tsne_combined}
\end{figure*}

To validate our approach, we initially trained VAE-MMD using data from two centers: Stanford and UCSF. Figure~\ref{fig:tsne_combined}(a) top illustrates the t-SNE visualization of latent features both before and after the VAE-MMD training process. Before domain adaptation (Fig. \ref{fig:tsne_combined}a top left), samples from Stanford (blue) and UCSF (red) exhibited distinct clusters with a noticeable separation. This indicates a significant domain gap between the two groups, attributable to differences in their lesion characteristics and imaging protocols. After training with VAE-MMD (Fig. ~\ref{fig:tsne_combined}a top right), the two distributions were much more similar, and samples from both institutions were mixed together throughout the feature space. This visual evidence suggests that VAE-MMD developed domain-invariant representations that preserved anatomically relevant information while removing biases specific to individual institutions.

\subsubsection{Four-Center Validation}

We broadened our evaluation to include all four centers—Stanford, UCSF, UCLM, and PKG—to assess the scalability of the domain adaptation methodology. Figure~\ref{fig:tsne_combined}a bottom shows t-SNE graphs of all the institutions before and after VAE-MMD training. Before domain adaptation (Fig. ~\ref{fig:tsne_combined}a bottom left), each institution established a distinct cluster. UCLM occupied a separate region, UCSF was positioned at the center, Stanford was located in its own area, and PKG was situated in a distinct zone. This distinct division indicated a significant transformation within the domain across all four centers.

After VAE-MMD training (Fig.~\ref{fig:tsne_combined}a bottom right), the clustering that was unique to each institution was significantly diminished. Samples from all four centers were combined, with no single institution overseeing any specific segment of the latent space. Even UCLM, which previously exhibited the most distinct separation prior to adaptation, became integrated into samples from other institutions.

\subsubsection{Domain Classifier Performance}

We developed a logistic regression classifier aimed at predicting institution labels based on latent features, with the goal of assessing the effectiveness of domain adaptation. Figure~\ref{fig:tsne_combined}(b) displays the confusion matrices prior to and subsequent to VAE-MMD training.

Before domain adaptation, the classifier was 91.0\% accurate, correctly identifying the institution of origin for most samples. The confusion matrix showed that Stanford, UCSF, and UCLM were almost perfectly separate, with only a little bit of confusion between Stanford and PKG. This high level of classification accuracy indicated that the original feature representations possessed distinct signatures specific to each institution, which could complicate efforts to generalize findings across different institutions.

After training with VAE-MMD, the accuracy of the classifier dropped to 50.0\%, which is close to the theoretical lower bound for a balanced four-class classification task (25\% random chance). The confusion matrix showed that the classifier now mostly predicted UCSF for samples from all institutions. This indicated that the model could no longer reliably tell the difference between centers based on latent features. This huge drop in the accuracy of the domain classifier, from 91.0\% to 50.0\%, corresponded to a 45.1\% reduction in domain separability, demonstrating that VAE-MMD successfully got rid of biases that were specific to each institution from the learned representations.

\subsection{Segmentation Performance}

We evaluated the performance of nnU-Net both with and without VAE-MMD preprocessing in configurations involving two, three, and four centers (Table~\ref{tab:segmentation}) to determine whether domain adaptation enhanced subsequent segmentation outcomes.

\begin{table*}[!t]
\centering
\caption{Comparison of segmentation performance between baseline mixed training and VAE-MMD preprocessing across multi-center configurations. Sens: sensitivity; Prec: precision; sDice: surface Dice; HD95: 95th percentile Hausdorff distance (mm).}
\label{tab:segmentation}
\begin{tabular}{llcccccccccccc}
\hline
 & & \multicolumn{6}{c}{Baseline} & \multicolumn{6}{c}{VAE-MMD} \\
\cline{3-8} \cline{9-14}
Configuration & Test & Sens & Prec & F1 & F2 & sDice & HD95 & Sens & Prec & F1 & F2 & sDice & HD95 \\
\hline
\multirow{2}{*}{Two-center} 
 & Stanford & 0.384 & 0.889 & 0.536 & 0.433 & 0.6831 & 12.78 & 0.652 & 0.683 & 0.638 & 0.649 & 0.7827 & 6.76 \\
 & UCSF & 0.756 & 0.894 & 0.819 & 0.780 & 0.8102 & 2.23 & 0.750 & 0.840 & 0.772 & 0.757 & 0.8251 & 2.07 \\
\hline
\multirow{3}{*}{Three-center} 
 & Stanford & 0.455 & 0.446 & 0.412 & 0.424 & 0.6738 & 14.79 & 0.628 & 0.684 & 0.635 & 0.626 & 0.6989 & 10.97 \\
 & UCSF & 0.767 & 0.866 & 0.798 & 0.776 & 0.8109 & 2.43 & 0.786 & 0.894 & 0.826 & 0.799 & 0.8200 & 2.44 \\
 & UCLM & 0.730 & 0.793 & 0.715 & 0.702 & 0.7358 & 4.43 & 0.850 & 0.791 & 0.792 & 0.808 & 0.8144 & 1.37 \\
\hline
\multirow{4}{*}{Four-center} 
 & Stanford & 0.568 & 0.587 & 0.563 & 0.563 & 0.6615 & 22.46 & 0.608 & 0.679 & 0.618 & 0.609 & 0.7278 & 2.12 \\
 & UCSF & 0.664 & 0.861 & 0.702 & 0.671 & 0.6782 & 7.26 & 0.780 & 0.832 & 0.792 & 0.782 & 0.7916 & 3.30 \\
 & UCLM & 0.843 & 0.830 & 0.805 & 0.817 & 0.8051 & 6.64 & 0.896 & 0.873 & 0.874 & 0.883 & 0.8382 & 2.99 \\
 & PKG & 0.701 & 0.804 & 0.728 & 0.709 & 0.7038 & 8.97 & 0.855 & 0.822 & 0.827 & 0.840 & 0.7168 & 7.21 \\
\hline
\end{tabular}
\end{table*}

\subsubsection{Two-Center Results}

In the two-center setup (Stanford and UCSF), VAE-MMD preprocessing made Stanford test results much better on all measures. Sensitivity increased from 0.384 to 0.652 (+0.268), and the F1 score increased from 0.536 to 0.638 (+0.102). This improvement was especially impressive because Stanford, which had a smaller dataset, was underrepresented in baseline mixed training. The substantial increase in sensitivity indicated that VAE-MMD helped the model to detect lesions that baseline models missed due to Stanford-specific imaging features. The quality of the surface boundary improved as well. Stanford's sDice increased from 0.6831 to 0.7827, while HD95 decreased from 12.78 mm to 6.76 mm. This indicates that the tumor boundaries were delineated with greater precision following domain adaptation. UCSF's performance stayed stable with an F1 score of 0.772. This demonstrates that domain adaptation improved generalization across institutions without hurting performance at well-represented centers.

\subsubsection{Three-Center Results}

Adding UCLM to the three-center configuration showed more consistent improvements across all centers. Stanford had the biggest improvements: sensitivity increased from 0.455 to 0.628 (+0.173), precision increased from 0.446 to 0.684 (+0.238), and F1 increased from 0.412 to 0.635 (+0.223). UCSF did a little better on all metrics, with F1 increasing from 0.798 to 0.826 (+0.028). UCLM's sensitivity increased from 0.730 to 0.850 (+0.120), and F1's increased from 0.715 to 0.792 (+0.077). At the same time, HD95 decreased from 4.43 mm to 1.37 mm, indicating that the boundaries were much more tightly agreed upon after adaptation. The mean F1 score across the three centers increased from 0.642 to 0.751, reflecting an enhancement of 0.109, which corresponds to a relative improvement of 17.0

\subsubsection{Four-Center Results}

\begin{figure*}[!t]
\centering
\includegraphics[width=\textwidth]{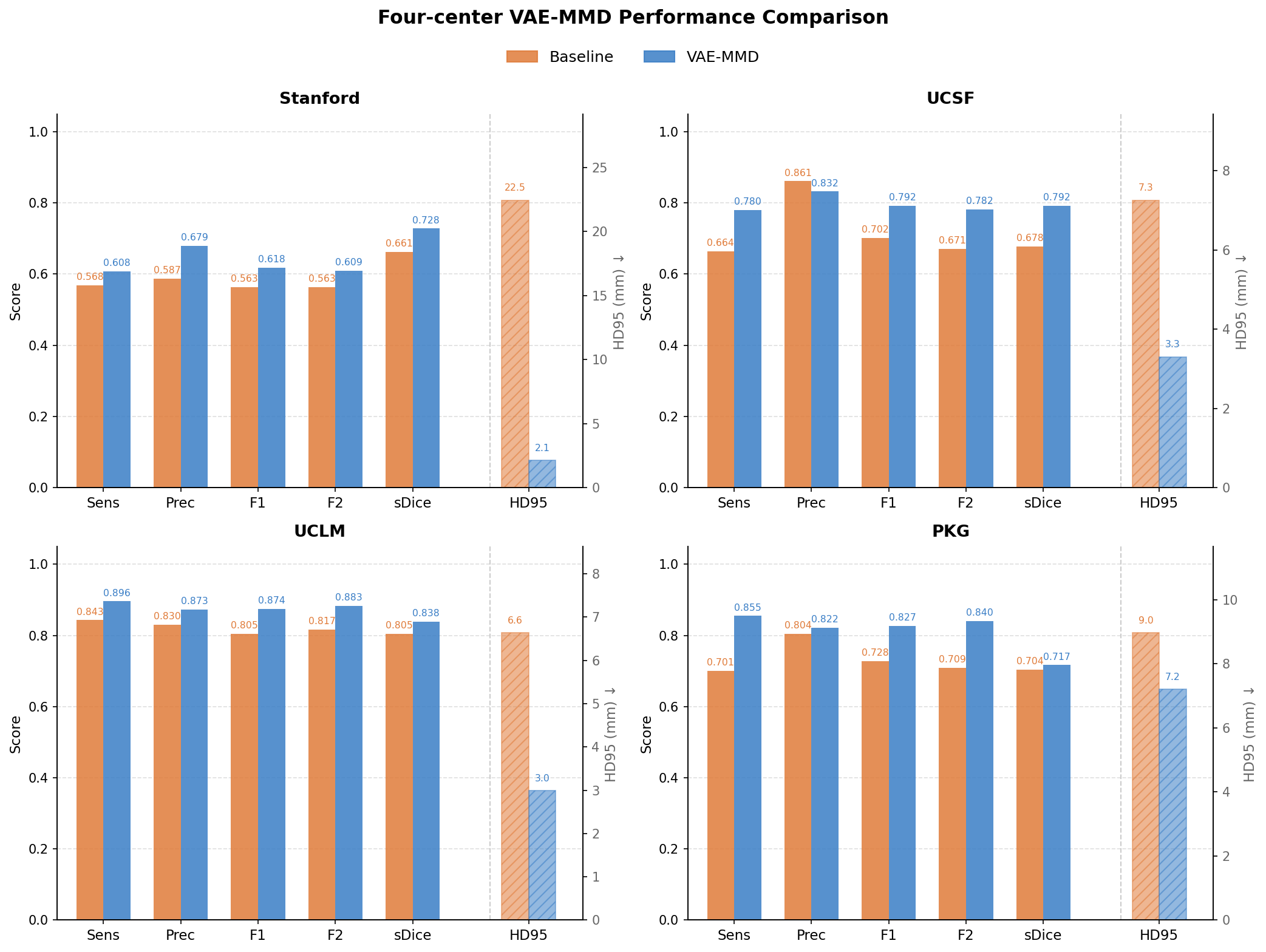}
\caption{Comparison of the performance of four-center segmentation prior to and following VAE-MMD preprocessing. The blue bars show performance after VAE-MMD, while the orange bars show baseline performance. VAE-MMD consistently raised sensitivity, F1, and surface Dice (sDice) scores at most institutions. Stanford, UCSF, and UCLM saw the biggest drops in HD95, which means that domain adaptation made it easier to draw boundaries.}
\label{fig:performance_comparison}
\end{figure*}

The four-center arrangement resulted in consistent advancements across all institutions. Stanford's score in F1 increased from 0.563 to 0.618 (+0.055), UCSF's score increased from 0.702 to 0.792 (+0.090), UCLM's score increased from 0.805 to 0.874 (+0.069), and PKG's score increased from 0.728 to 0.827 (+0.099). The sensitivity improvements were particularly pronounced for PKG (+0.154), indicating that VAE-MMD preprocessing helped detect lesions that baseline models missed in this dataset with uniform primary tumor histology. The average F1 score increased from 0.700 to 0.778, reflecting a rise of 0.078, which corresponds to an 11.1\% relative enhancement across all four centers. The most substantial changes in surface metrics were observed at Stanford, where HD95 dropped dramatically from 22.46 mm to 2.12 mm. This finding is consistent with the fact that small metastases are common in that dataset, where baseline models had high miss rates that artificially inflated surface distances. UCLM and UCSF also had consistent HD95 reductions (6.64→2.99~mm and 7.26→3.30~mm, respectively), while sDice improved across Stanford, UCSF, and UCLM. PKG had a small sDice improvement (0.7038→0.7168) and a smaller HD95 reduction (8.97→7.21~mm), which suggests that this dataset had less boundary-level gains despite strong volumetric improvements. UCLM achieved the best absolute performance (F1=0.874, F2=0.883) after VAE-MMD preprocessing, even though it showed the most pronounced domain shift in the t-SNE visualization before adaptation. This indicates that our method capable of bridging even substantial domain gaps. Figure~\ref{fig:performance_comparison} illustrates the improvements achieved across all metrics and in every center as a result of these changes.

\begin{figure*}[!t]
\centering
\begin{tabular}{cc}
\multicolumn{1}{c}{\textbf{Baseline}} & \multicolumn{1}{c}{\textbf{After VAE-MMD}} \\[0.3em]
\includegraphics[width=0.48\textwidth]{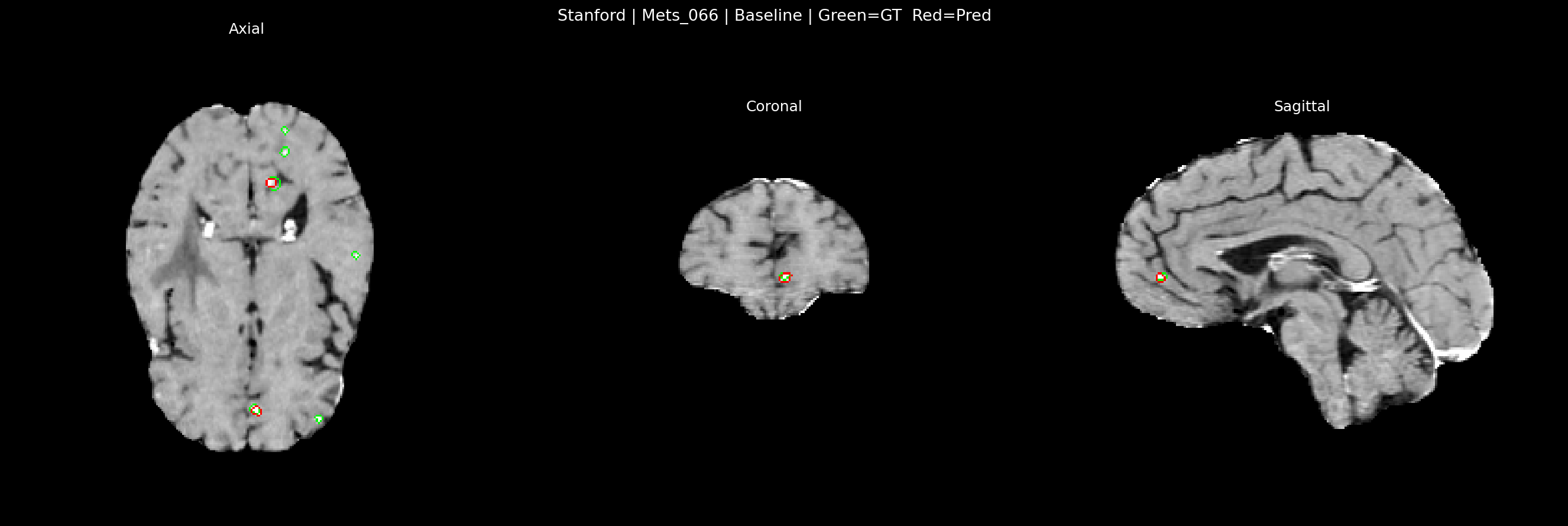} &
\includegraphics[width=0.48\textwidth]{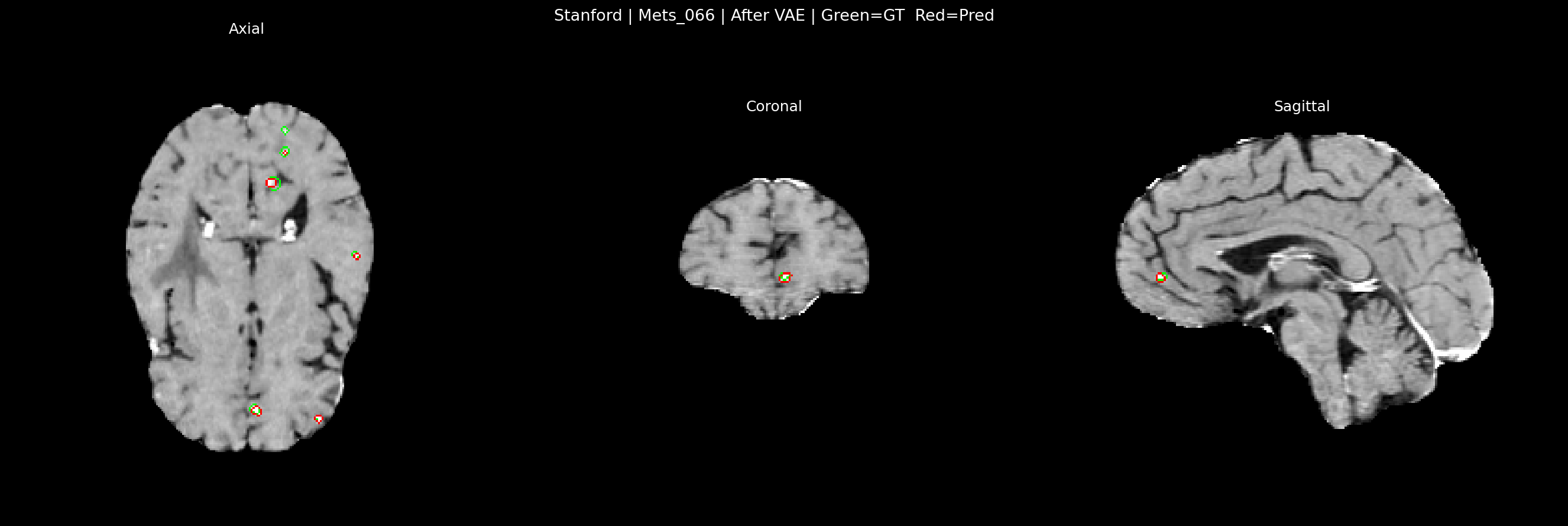} \\[0.2em]
\includegraphics[width=0.48\textwidth]{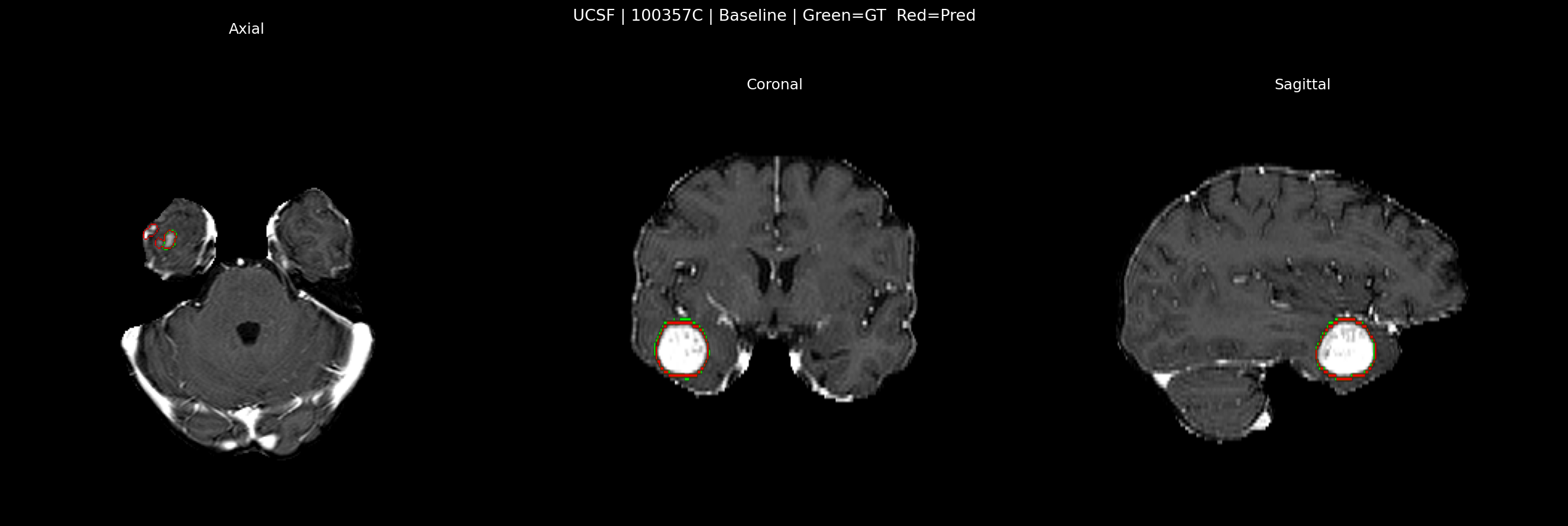} &
\includegraphics[width=0.48\textwidth]{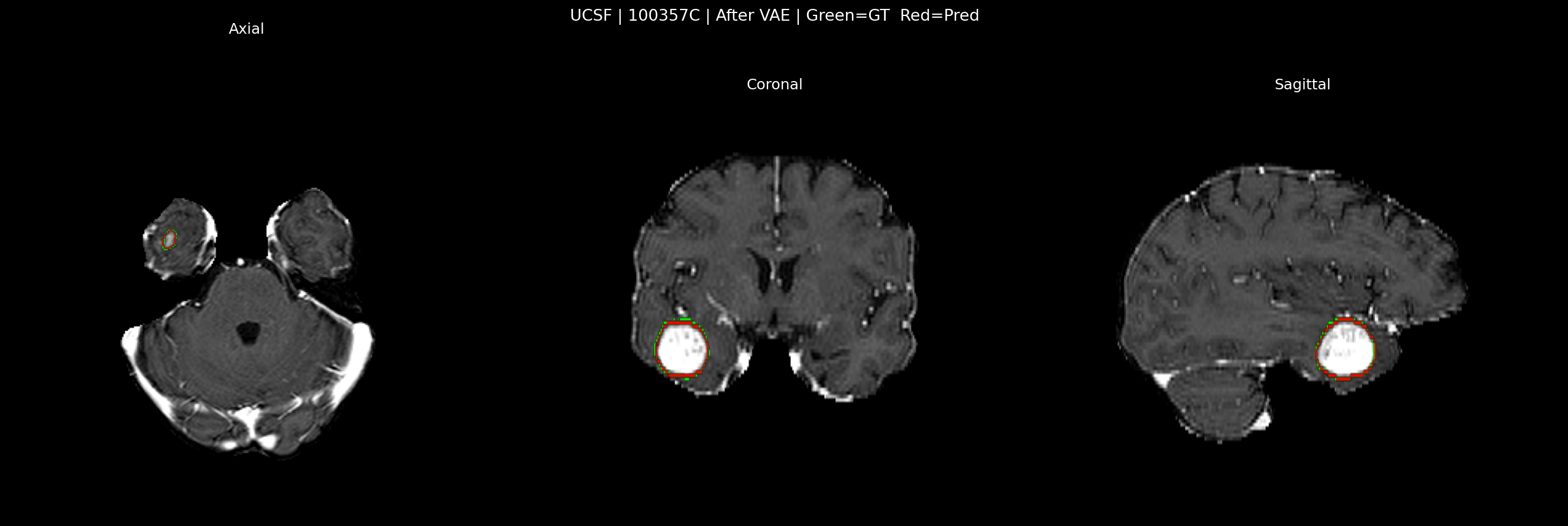} \\[0.2em]
\includegraphics[width=0.48\textwidth]{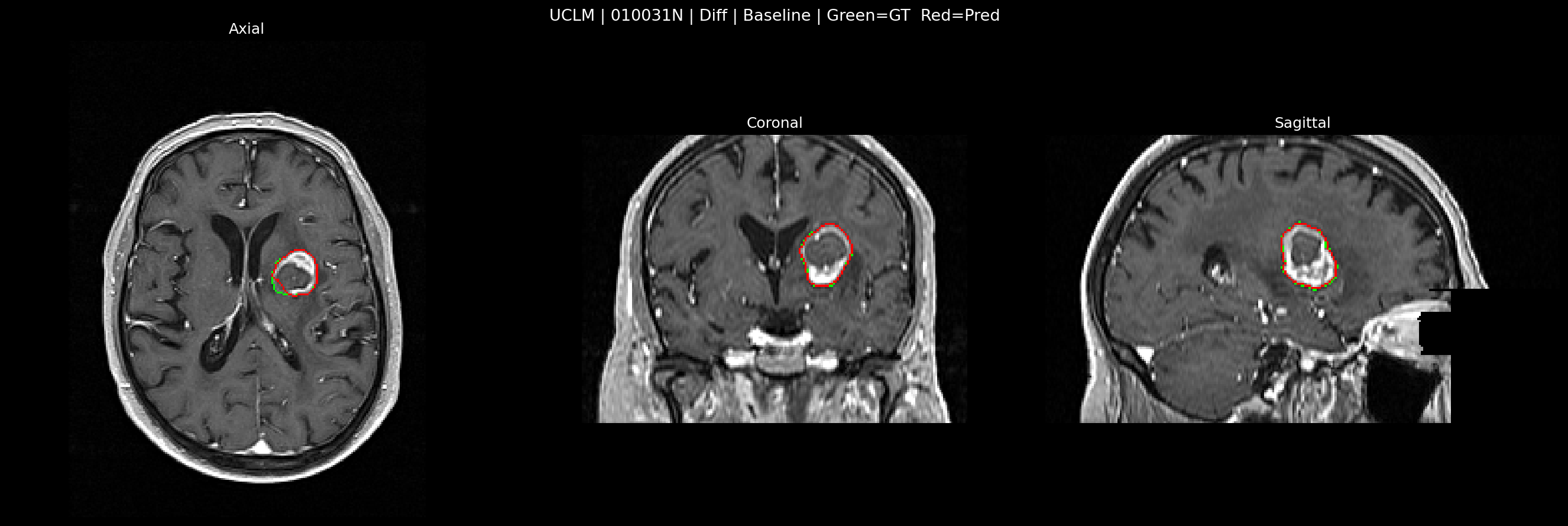} &
\includegraphics[width=0.48\textwidth]{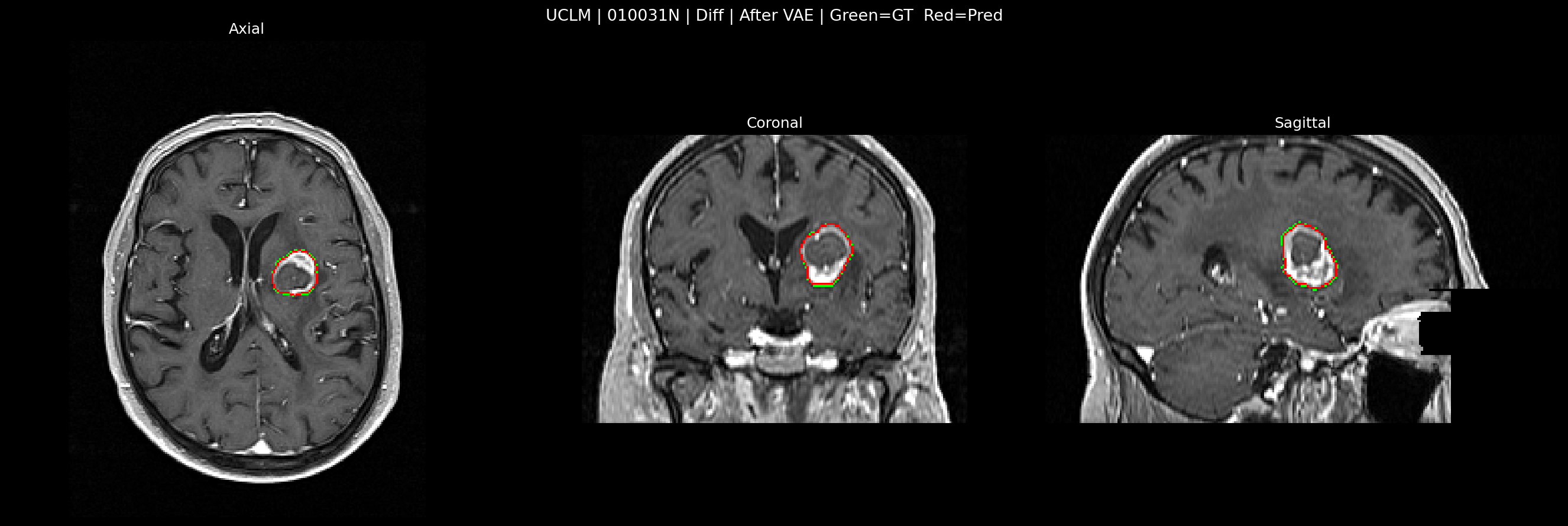} \\[0.2em]
\includegraphics[width=0.48\textwidth]{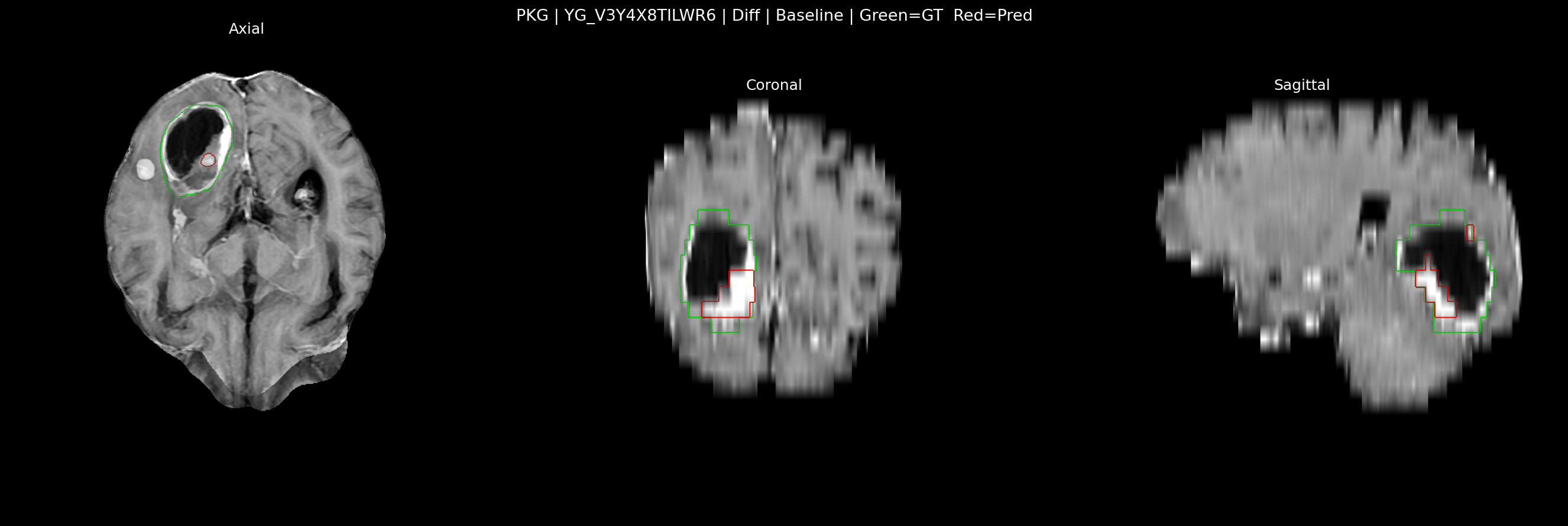} &
\includegraphics[width=0.48\textwidth]{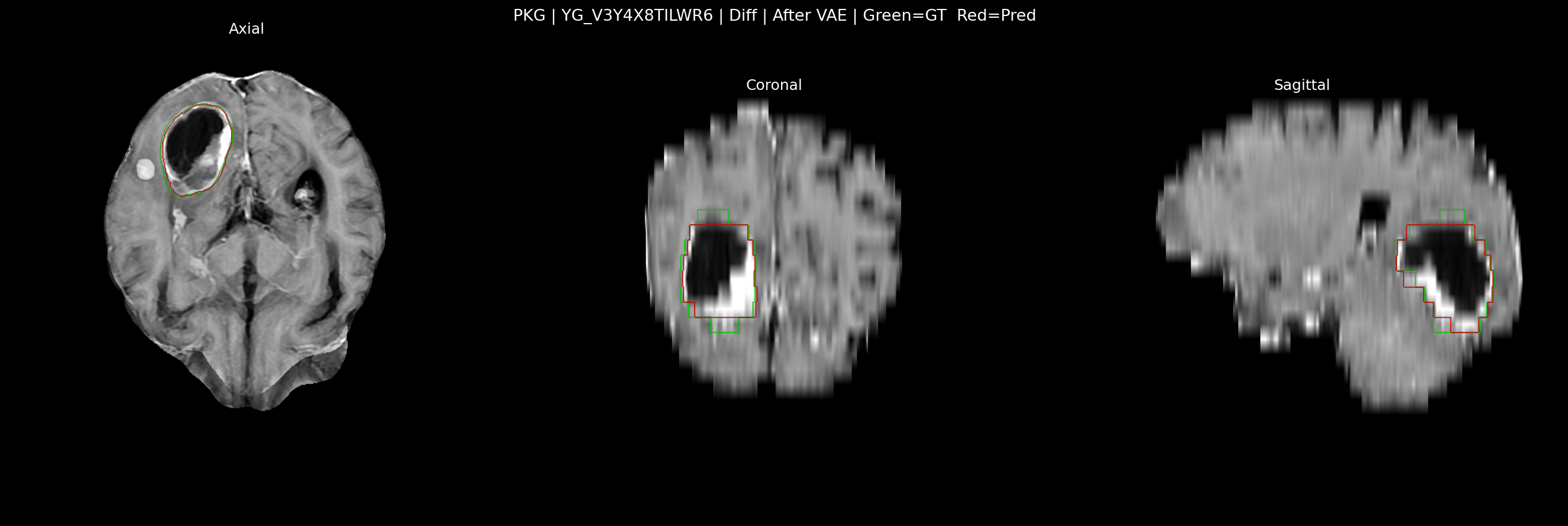} \\
\end{tabular}
\caption{Qualitative comparison of segmentation results before (left, Baseline) and after (right, VAE-MMD) domain adaptation across four datasets (Stanford, UCSF, UCLM, PKG from top to bottom). Green contours indicate ground truth annotations; red contours indicate model predictions. Please zoom in the digital manuscript for better visualization.}
\label{fig:qualitative}
\end{figure*}

\section{Discussion}

This study demonstrated that VAE-MMD domain adaptation successfully lessened cross-institutional domain shift and enhanced BM segmentation performance across various centers. Our method resulted in a 45.1\% relative decrease in domain classifier accuracy (from 91.0\% to 50.0\%), an 11.1\% relative increase in mean F1 score, a 7.93\% improvement in mean sDice (0.7121$\to$0.7686), and a 65.5\% reduction in mean HD95 (11.33$\to$3.91 mm) across four institutions. These surface-level gains are particularly noteworthy: HD95 reductions were observed at all four centers---most dramatically at Stanford (22.46$\to$2.12 mm)---and sDice improved consistently at three of four centers. Taken together, these results confirm that VAE-MMD improved not only volumetric lesion detection but also the geometric accuracy of boundary delineation. These findings address a major challenge in the clinical application of AI-assisted segmentation tools.

\subsection{Effectiveness of VAE-MMD for Domain Adaptation}

Several important design choices contributed to the success of VAE-MMD. First, MMD alignment operated in a learned latent space rather than pixel space, allowing it to focus on high‑level semantic features. It ignored low-level domain-specific variations such as noise patterns and intensity distributions. This method kept anatomically important information that was necessary for segmentation while getting rid of biases that were specific to the institution. Second, the multi-scale RBF kernels ($\sigma \in \{0.5, 1.0, 2.0\}$) captured domain discrepancy at different levels of detail, allowing for the alignment of both fine-grained and global distributional differences at the same time. Third, skip connections in the decoder architecture kept high-frequency spatial details that would have been lost through the information bottleneck. This made sure that the edges of the lesions stayed sharp after reconstruction. The improvements in surface-level metrics, especially the substantial HD95 drops observed at Stanford, UCSF, and UCLM, suggest that this boundary preservation also applied to the downstream segmentation task, resulting in more geometrically accurate predictions aligned with ground truth annotations.

The t-SNE visualizations provided strong evidence that the domain alignment was effective. Before VAE-MMD, each institution formed distinct clusters in feature space, with UCLM exhibiting the most pronounced separation. After training, samples from all four centers mixed together in the latent space, indicating that the learned representations no longer contained information that was specific to each institution. This alignment did not compromise anatomical information, as evidenced by the better segmentation performance across all volumetric metrics and surface-based measures, including sDice and HD95.

\subsection{Comparison with Other Approaches}

Our results outperformed those of other domain adaptation techniques reported in the literature. Our previous research demonstrated that naive transfer learning suffers from catastrophic forgetting, as shown by a drop in precision from 0.900 to 0.418 when models were transferred between centers.\cite{huang2024multicenter} Conversely, VAE-MMD maintained or enhanced accuracy across the majority of configurations while significantly boosting sensitivity. Moreover, our method works in an unsupervised way with target domains, so it doesn't need any labeled data from new institutions. This represents a significant advantage in clinical settings.

The observation that UCLM achieved the best absolute performance (F1=0.874) after VAE-MMD preprocessing, even though it had the most substantial domain shift before adaptation, suggests that our method is particularly effective at bridging big domain gaps. This discovery carries important consequences for the application of models in institutions that display imaging features significantly different from those in the training data.

\subsection{Clinical Implications}

The improvements observed in this study are directly useful for planning stereotactic radiosurgery. The significant increases in sensitivity, especially for PKG (+0.154) and Stanford (+0.268 in the two-center configuration), demonstrate that VAE-MMD preprocessing helps detect lesions that baseline models miss. Failure to detect lesions during treatment planning can exacerbate the disease and negatively impact the patient's condition. Therefore, high sensitivity in clinical applications is crucial.

The reductions in HD95 observed following VAE-MMD preprocessing hold significant importance for clinical applications. In the four-center configuration, the Stanford HD95 measurement decreased from 22.46 mm to 2.12 mm, indicating a transition from significant mislocalization of the boundaries to a clinically acceptable level. In radiosurgery planning, where treatment margins are usually between 1 and 3 mm, accurately defining the edges of the tumor and the healthy tissue around it directly affects how much radiation is delivered to the tumor and how much healthy tissue is spared. The consistent enhancements in sDice indicate that VAE-MMD achieved segmentations with improved surface overlap, which is more closely tied to the precision of treatment volume than volumetric overlap by itself.

Our approach addresses a significant challenge that has hindered the implementation of AI-assisted segmentation in hospitals: the necessity to retrain or fine-tune models for every new healthcare facility. With VAE-MMD preprocessing, one model trained on data from multiple institutions can work with new centers without needing local labeled data or changes to the model. This capability has the potential to accelerate the adoption of AI tools in radiation oncology, particularly in institutions that lack the financial resources for extensive local validation or model development.

\subsection{Limitations}

It is essential to acknowledge certain limitations. Initially, we relied exclusively on T1CE sequences to maintain uniformity across different centers, even though FLAIR and other sequences could provide additional insights for identifying lesions. Future research could explore multi-modal VAE architectures that simultaneously align different imaging sequences. Second, VAE reconstruction naturally entails a degree of information loss via the latent bottleneck, although our skip connection architecture alleviated this phenomenon. Third, our approach does not require labeled data from the target domain; however, it necessitates access to unlabeled images from multiple institutions during the training of the Variational Autoencoder (VAE). This requirement may pose challenges in certain situations. Fourth, we utilized retrospective data with standardized train/test splits to assess our findings. Validation on genuinely external institutions would provide more robust evidence of its effectiveness in diverse settings. Fifth, surface metrics such as sDice and HD95 were not accessible for every configuration due to the absence of saved predictions from previous experiments. These metrics were calculated solely for the four-center configuration in baseline models and for all configurations in VAE-MMD models. Future research should ensure that surface metrics are consistently reported across all experimental conditions.

\subsection{Future Directions}

There are a number of ways this work could be expanded. Federated learning frameworks would enable multiple institutions train a VAE-MMD without centralizing patient data, addressing privacy concerns that complicate data sharing.\cite{huang2024multicenter} Exploring different strategies for domain adaptation, such as adversarial domain adaptation or optimal transport methods, could lead to better alignment results. Additionally, expanding the framework to encompass additional brain tumor segmentation tasks, such as glioma or meningioma, would illustrate its wider applicability. Finally, prospective clinical validation studies comparing AI-assisted segmentation with manual segmentation, both with and without domain adaptation. This approach will provide us with definitive evidence regarding the effectiveness of these methods in real-world applications.

\section{Conclusion}

We proposed a VAE-MMD framework designed for cross-institutional domain adaptation in brain tumor segmentation. Our method combined representation learning based on variational autoencoders with domain alignment based on maximum mean discrepancy. This reduced domain classifier accuracy from 91.0\% to 50.0\%, improved mean F1 by 11.1\% (0.700$\to$0.778), mean sDice by 7.93\% (0.7121$\to$0.7686), and reduced mean HD95 by 65.5\% (11.33$\to$3.91 mm) across four institutions (Stanford, UCSF, UCLM, and PKG). These boundary-level improvements---HD95 reductions at all four centers and sDice gains at three of four centers---confirm that VAE-MMD enhanced geometric precision of boundary delineation beyond volumetric detection alone. It doesn't need any labeled data from the target domains, and it can work with existing segmentation frameworks like nnU-Net. These results show that VAE-MMD effectively solves the domain shift problem that has made it difficult to use AI-assisted BM segmentation in clinical settings. This could lead to more widespread use of automated tools for planning stereotactic radiosurgery.

\section*{Funding}

This work was supported by Peking University Third Hospital Beijing Key Laboratory of Intelligent Neuromodulation and Brain Disorder Treatment.

\section*{Conflict of Interest}

The authors declare no conflict of interest.

\section*{Authorship}

Y.Y. conducted experiments and analyzed data. Y.Y. and Y.H. designed the experimental framework, drafted and revised the manuscript. Y.H. and F.P. conceived the study and developed the hypothesis. S.Z. revised the manuscript. S.Z., H.Y., L.S. collected data. H.H. provided technical resources. Y.H. supervised the study. All authors read and approved the final version of the manuscript.

\section*{Data Availability}

The implementation code and links to the publicly available BM datasets can be found at \url{https://github.com/BigMewin/BM}.

\section*{Ethics}

All datasets used in this study are publicly available. The data collection was approved by the Institutional Review Boards (IRB) of the respective contributing institutions, and data sharing was performed in accordance with their Data Transfer Agreements (DTAs). The data were fully anonymized prior to release, and therefore, this retrospective analysis did not require additional ethical approval from our institution.

\bibliographystyle{unsrt}
\bibliography{references}

\end{document}